# Prompt-RAG: Pioneering Vector Embedding-Free Retrieval-Augmented Generation in Niche Domains, Exemplified by Korean Medicine

*Bongsu Kang[1], Jundong Kim[1], Tae-Rim Yun[1], Chang-Eop Kim[1, 2, *]*

[1]Department of Physiology, College of Korean Medicine, Gachon University, Seongnam, Gyeonggi, Republic of Korea
[2]Department of Neurobiology, Stanford University School of Medicine, Stanford, California, USA

\* Corresponding Author: Chang-Eop Kim
  Email: eopchang@gachon.ac.kr

## ABSTRACT

We propose a natural language prompt-based retrieval augmented generation (Prompt-RAG), a novel approach to enhance the performance of generative large language models (LLMs) in niche domains. Conventional RAG methods mostly require vector embeddings, yet the suitability of generic LLM-based embedding representations for specialized domains remains uncertain. To explore and exemplify this point, we compared vector embeddings from Korean Medicine (KM) and Conventional Medicine (CM) documents, finding that KM document embeddings correlated more with token overlaps and less with human-assessed document relatedness, in contrast to CM embeddings. Prompt-RAG, distinct from conventional RAG models, operates without the need for embedding vectors. Its performance was assessed through a Question-Answering (QA) chatbot application, where responses were evaluated for relevance, readability, and informativeness. The results showed that Prompt-RAG outperformed existing models, including ChatGPT and conventional vector embedding-based RAGs, in terms of relevance and informativeness. Despite challenges like content structuring and response latency, the advancements in LLMs are expected to encourage the use of Prompt-RAG, making it a promising tool for other domains in need of RAG methods.

## 1. Introduction

Retrieval-Augmented Generation (RAG) models combine a generative model with an information retrieval function, designed to overcome the inherent constraints of generative models.(1) They integrate the robustness of a large language model (LLM) with the relevance and up-to-dateness of external information sources, resulting in responses that are not only natural and human-like but also the latest, accurate, and contextually relevant to the query.(1-4) The interaction of the two modules (retrieval and generation) enables responses that would not be achievable with either module alone, making RAG more than just the sum of its components. This approach represents a significant milestone in the field of generative models by enabling the induction of high-quality responses in less-explored domains at a low expense.(5, 6)

In the conventional RAG operation, the initial step involves converting input queries into vector embeddings, which are then used to retrieve relevant data from the vectorized database. Following this, the generative part of RAG utilizes the retrieved external data for producing contextually rich responses.(7) Thus, both the embedding and generative models are considered crucial factors in the performance of RAG, directly affecting the retrieval process.(8) However, in niche domains, the performance of generic LLM-based embedding models appears suboptimal compared to their effectiveness in more general fields. The lack of specialized training data in these domains results in embeddings that do not adequately capture the nuances and specificity of the domain(9), leading to less accurate and contextually relevant information retrieval. Despite the evident presence of these functional limitations, they have not been much identified through experiments, therefore the optimality of the conventional LLM-based vector embedding RAG methods for niche domains has remained in obscurity. Researchers have been aware of these shortcomings of LLMs and have explored supplementary processes such as fine-tuning to improve the performance.(8, 10-12) However, the cost of fine-tuning, especially when it involves adjusting the entire or majority of parameters in LLM, has rapidly become expensive, thereby increasing the demand for alternative solutions.(13-15)

To address these challenges, we propose a novel methodology: Prompt-RAG. This new approach to RAG eliminates the reliance on vector embeddings, adopting a more direct and flexible retrieval process based on natural language prompts. It involves a large-scale pre-trained generative model that handles the entire steps from document retrieval to response generation without the need for a vector database or an algorithm for indexing and selecting vectors, thus having the processing structure of RAG greatly simplified. Therefore, it not only takes advantage of the RAG's strength but also circumvents the limitations of conventional vector embedding-based methodology. Prompt-RAG is based on maximizing the use of the advanced natural language processing capabilities of LLMs. Especially using the latest GPT model, our method can compensate for the deficiencies in vector embedding-based RAG arising from the shortage of domain-specific knowledge.

To examine the utility of Prompt-RAG in practice, we conducted two exemplary studies focusing on the Korean Medicine (KM) domain. KM, a branch of traditional East Asian medicine, has diverged from traditional Chinese medicine and Japanese Kampo medicine in aspects like physiological theories, treatments, and Sasang constitutional medicine.(16, 17) It was reported that GPT models have achieved excellent results in the United States Medical Licensing Examination (USMLE)(18-20), while ChatGPT's scores on the Korean National Licensing Examination for Korean Medicine Doctors barely reached the passing threshold, underperforming in subjects unique to KM, especially Sasang constitutional medicine and public health & medicine-related law.(21) In this niche area, rich in specialized knowledge and distinct from Conventional Medicine (CM), we first demonstrated the functional suboptimality of LLM-based vector embeddings. Subsequently, we demonstrated Prompt-RAG's effectiveness in this context. A Question-Answering (QA) chatbot based on Prompt-RAG was built using KM-specific documents, and our model's performance was compared with that of ChatGPT and conventional vector embedding-based RAG models. This study not only highlights the challenges of conventional RAG methods in niche domains but also showcases the potential of Prompt-RAG as a more effective alternative.

## 2. Design of Prompt-RAG

In this study, we introduce Prompt-RAG, a novel approach distinct from the conventional vector embedding-based RAG. Prompt-RAG consists of three steps: preprocessing, heading selection, and retrieval-augmented generation. The overall scheme of Prompt-RAG might seem similar to that of conventional RAG methods. However, details in each step are quite distinguishable especially in that conventional RAGs rely on a complex multi-step process involving the vectorization of documents and algorithmic retrieval from a vector database for a generative model's response. The workflows of vector embedding-based RAG and our method are depicted in Figure 1.

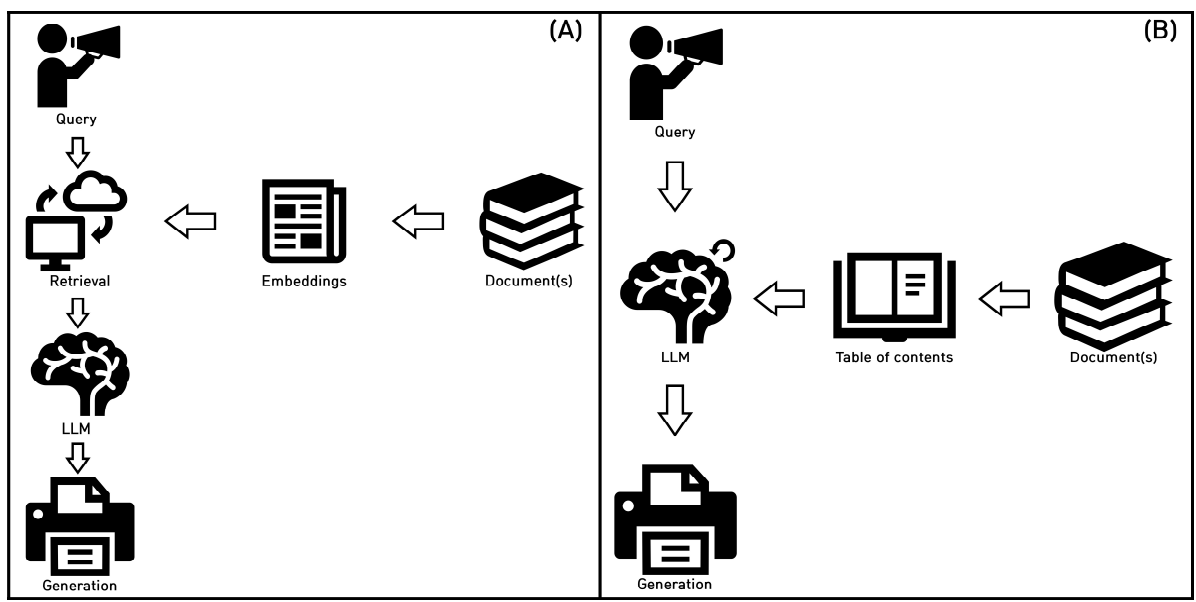

Figure. 1. Comparative workflows of two RAG models. (A) depicts the vector embedding-based RAG process. Relevant pieces of information are retrieved from a database of document embeddings through algorithms. The retrieved data are augmented in a generative model to produce a response. (B) illustrates the process of Prompt-RAG. An LLM-based generative model directly uses a table of contents for constructing a contextual reference, followed by generating a response with it.

Abbreviation: RAG, Retrieval-augmented generation; LLM, Large-language model.

1) Preprocessing

Prompt-RAG initiates by extracting or creating a Table of Contents (ToC) from a user's document(s), which is the main subject of the retrieval. The procedure can be done flexibly depending on the type of document and the user's preferences. One of the most ideal cases is that a ToC is already prepared, made by the author(s) of the document. And yet, even in the absence of a pre-determined ToC, it can be arbitrarily generated, for example, using a generative model or in a manual way, based on the document's quantitative, semantic, or individual divisions. It should be noted that the size of a ToC must not exceed the context window size of the generative model for heading selection. Consequently, some headings or details of the ToC (e.g., heading or page numbers, or hierarchical structure) might need to be removed in order to reduce the number of tokens. The body of the document should then be divided into sections according to the headings and prepared for subsequent retrieval.

2) Heading selection

A prompt, which contains both a query and a ToC, is passed to an LLM-based generative model and the model is asked to autonomously select the headings most pertinent to the query or those that help the most to find information concerning the query. Multiple heading selections can be performed using the hierarchical structure of the headings, narrowing down from main headings to subheadings if a user wants to make use of all the headings from an oversized ToC. As this procedure is a preliminary step for making a reference for answer generation, the number of selected headings can be set in the prompt in advance depending on the budget and the context window size of the generative model for answer generation. It is recommended that the model produce a response in a structured format during heading selection to optimize efficiency for the following retrieval process as well as token usage.

3) Retrieval-augmented generation

Sections of the document under the selected headings are retrieved and concatenated as a reference for answer generation. Again, it should be noted that the size of a reference must be smaller than the context window size of the generative model for answer generation. Therefore, the size of a reference has to be reduced by truncation or summarization when overly large. After a reference is prepared, a prompt including both the query and the reference is forwarded into a generative model. In response, the model consults the augmentations to generate a response to the query.

## 3. Experiments

1) Comparative exploration of LLM-based vector embeddings in the KM and CM domains.

This experiment aimed to identify and exemplify the relative representational defects of LLM-based vector embedding in niche domains compared to other well-established domains. To explain this point, we conducted a comparative analysis with vector embeddings from documents in KM and CM domains.

For this experiment, we selected 10 documents each from KM and CM domains, specifically regarding their physiological contents. '*Eastern Medicine Physiology*'(22) served as the document pool for KM. This book, compiled in Korean, has been revised by professors from every Korean Medicine college in South Korea and is used as the principal textbook in the physiology curriculum. On the other hand, '*Physiology*'(23) was chosen for the CM domain. To investigate the impact of language on representational differences in embeddings, we collected documents with the exactly identical contents from both the English version and the Korean-translated version of '*Physiology*'. The titles of the selected documents from each domain are listed in Appendix Table 1. We extracted the embedding vectors for a total of 30 documents – 10 each from KM physiology, CM physiology in Korean (CM_KR), and CM physiology in English (CM_EN) – using E5-mistral-7b-instruct(24), Voyage AI's voyage-02, and OpenAI's text-embedding-ada-002 models to figure out LLMs' representations of KM and CM knowledge.

Our analysis focused on identifying patterns of the KM and the CM domain embeddings with three key document similarity metrics: human-evaluated document relatedness, embedding correlation coefficients, and token overlap coefficients. We assessed whether the correlation coefficients between embedding pairs closely align with the human-evaluated ground truth or merely follow the surface-level similarity (token overlap) by conducting the correlation analyses across these metrics. It allows us to understand the depth of embedding representations and their correlation with human-perceived document pairwise relevance.

For this, the Pearson correlation coefficients(25) were calculated for every embedding vector pair, covering 45 pairs in each of the three categories (KM, CM_KR, CM_EN). To assess explicit similarity in a document pair, we computed the overlap coefficient(26) for tokens in KM, CM_KR, CM_EN documents. The token overlap coefficient was calculated as:

$$Token\ overlap\ coefficient_{A,B} = \frac{|\,A \cap B\,|}{\min(|A|, |B|)}$$

$|\,A \cap B\,|$: The count of token co-occurrence between documents A and B.

$\min(|A|, |B|)$: The minimum token count in either document A or B.

Token overlap coefficients were calculated three times with different tokenizers corresponding to the embedding models: E5-mistral-7b-instruct(24), Voyage AI's voyage-02, and OpenAI's text-embedding-ada-002. Repeated appearances of a single token in a document were counted and considered separately.

To determine the ground truth of document pair correlations within each domain, two KM doctors with national licenses evaluated the relatedness between each pair of the KM and CM documents. A binary scoring system was adopted: a score of 1 indicated that a pair was interrelated, and 0 for unrelated documents. The human-evaluated document relatedness scores were then obtained by averaging the two doctors' scores in KM and CM documents, respectively.

The correlation analyses were conducted between human-evaluated document relatedness scores and embedding correlation coefficients, and between embedding correlation coefficients and token overlap coefficients with Scipy(27) in Python 3.11. Bonferroni correction(28) was applied for p-values due to the multiple comparisons.

2) Performance comparison of Prompt-RAG and existing models

(1) Chatbot Settings

For the evaluation, we developed a domain-specific, prompt-RAG-based chatbot for the book *'Introduction to Current Korean Medicine'*(29). The chatbot employed GPT architectures: GPT-4-0613 for the heading selection and GPT-3.5-turbo-16k-0613 for the answer generation.

The original ToC of the book had already been defined by the authors. Subheadings were added to it, aligning with the book's actual sections. The expanded table of contents exceeded the context window size for heading selection, so some headings were removed to handle this issue. The body of the book was then segmented according to the modified headings for the subsequent retrieval.

We passed a model based on GPT-4 a prompt containing both the revised ToC and a query, asking the model to identify five pertinent headings from the ToC. At the same time, it was instructed to avoid selecting a heading if the query was about greetings or casual talks. The prompt for heading selection is shown in Table 1.

Table 1. The prompt for heading selection

"Current context:
{history}[a]

Question: {question}[a]

Table of Contents:
{index}[a]

Each heading (or line) in the table of contents above represents a fraction in a document.
Select the five headings that help the best to find out the information for the question.
List the headings in the order of importance and in the format of
'1. ---
2. ---
---
5. ---'.
Don't say anything other than the format.
If the question is about greetings or casual talks, just say 'Disregard the reference.'."

[a]These represent the placeholders for conversational buffer memory, the user's query, and the table of

contents, respectively, from top to bottom.

Upon selecting the headings, the corresponding book sections were fetched and concatenated. In turn, this was provided as a reference in a prompt along with the query to another generative model based on GPT-3.5-turbo-16k. This model was required to generate an answer with the prompt which also contained a directive to refrain from saying nonsense when no relevant context was found in the reference thereby aiming to minimize hallucination. In cases where the selected headings are absent due to the query being a greeting or casual conversation, an alternative prompt without a reference section is passed to a GPT-3.5-turbo-based model, in order to reduce token usage and save on expenses. The prompts for answer generation are depicted in Table 2.

Table 2. The prompts for answer generation

| Prompt 1: Answer generation with selected headings |
| --- |
| "You are a chatbot based on a book called '현대한의학개론'.<br><br>Here is a record of previous conversation for your smooth chats.:<br>{history}[a]<br><br><br><br>Reference:<br>{context}[a]<br><br><br><br>Question: {question}[a]<br><br><br><br>Use the reference to answer the question.<br>The reference above is only fractions of '현대한의학개론'.<br>Be informative, gentle, and formal.<br>If you can't answer the question with the reference, just say like 'I couldn't find the right answer this |

time'.

Answer in Korean:"

---

**Prompt 2: Answer generation without selected headings for casual queries**

"You are a chatbot based on a book called '현대한의학개론'.

Here is a record of previous conversation for your smooth chats.:

{history}[a]

Question: {question}[a]

Answer the question.

Be informative, gentle, and formal.

Answer in Korean:"

---

[a]These denote the placeholders for conversational buffer memory, the reference based on the selected heading, and the user's query, respectively, from top to bottom.

Conversation buffer memory was incorporated in the prompts for both heading selection and answer generation, within each context window limit. We employed Langchain(30) for the processes above.

(2) Baselines

① ChatGPT

For the first baseline to compare the performance of our model with, we utilized ChatGPT without any retrieval-augmentation process. ChatGPT is based on a diverse, large-scale corpus, equipped with an immense range of global knowledge.(31) Therefore, we evaluated our model's proficiency in generating answers specific to the domain of KM, in contrast with general knowledge of ChatGPT. This baseline included employing both GPT-3.5 and GPT-4 models of ChatGPT (chatGPT-3.5, ChatGPT-4, respectively).

② Chunk retrievals

As our second baseline, we adopted vector embedding-based chunk retrieval. The text of the book was divided into chunks of size 50 and 100, respectively, using Tiktoken(32). Subsequently, each chunk was vectorized through OpenAI's text-embedding-ada-002. Vectors that most closely matched the query embedding by maximal marginal relevance(33) were retrieved. The number of retrieved vectors was set to 300 for chunk size 50 (C50-V300) and 150 for chunk size 100 (C100-V150), respectively, to make the most of the context window of GPT-3.5-turbo-16k for answer generation.

(3) Tasks and performance evaluation metrics

To evaluate the performance of our domain-specific, prompt-RAG-based chatbot and the other baseline models, we composed a series of 30 questions related to KM. The models were to generate answers to those questions in order.

Each question was categorized into one of the three types to examine the models' capabilities in direct retrieval, comprehensive understanding, and functional robustness. The questions among the three types followed a ratio of 4:4:2. For the ChatGPT baselines, which do not utilize retrieval augmentation, questions specifically inquiring about the author's perspective were appropriately adjusted. Further details on the questions and their types are provided in Appendix Table 2.

Human evaluation was performed for the generated answers by three KM doctors. The evaluators assessed the models' answers in terms of three criteria: relevance, readability, and informativeness.(34, 35) Relevance measured how well the answer directly addressed the central topic of the question. Readability evaluated the naturalness and fluency of the answer. Informativeness assessed the depth and significance of the answer's content. Each question was scored in terms of every criterion with either 0, 1, or 2 points. In the evaluation process, each response started with a base score of 2 for each criterion, and evaluators were instructed to deduct points based on the presence of specific flaws. Descriptions for the criteria and the scoring system are provided in Table 3. The Response time taken to generate each answer was also measured for the comparison of our model and chunk retrieval models

Table 3. Evaluation criteria for answers.

| Criterion | Point scale | Description | Deduction |
|---|---|---|---|
| **Relevance** | 0, 1, 2 | Assesses direct connection with the central topic of the question. High relevance achievable even with low readability or meaningless content. | Irrelevance to the question. |
| **Readability** | 0, 1, 2 | Evaluates the naturalness and fluency of an answer. High readability achievable even with irrelevant or meaningless content. | Grammatical errors or incoherence. |
| **Informativeness** | 0, 1, 2 | Assesses the depth and significance of the answer's content. High informativeness achievable even with low readability or irrelevance. | Superficial or meaningless content including hallucination. |
| **Scoring guide** | 0 points | Criterion severely damaged, making the answer unacceptable. | |

| | 1 point | Some flaws present in criterion, answer still usable. |
| | 2 points | Good overall criterion quality. |

(4) Statistical analysis

To evaluate the statistical significance of our model's scores in relation to those of the others, we performed t-tests and Mann-Whitney U tests. The t-tests compared the scores across the criteria of relevance, readability, and informativeness, while Mann-Whitney U tests were applied to the scores categorized by question types. P-values were adjusted using Bonferroni correction(28) to account for the multiple comparisons. All statistical analyses were conducted with the Statsmodels(36) package in Python 3.11.

## 4. Results

1) Comparative analysis of LLM-based vector embeddings in KM and CM

(1) Comparison of KM and CM document pairs by correlation metrics

Human-evaluated document relatedness scores, embedding correlation coefficients, and token overlap coefficients were calculated for KM and CM document pairs using three different embedding models. To compare the overall pattern of these metrics across the domains and the models, they are visually presented in Figure 2.

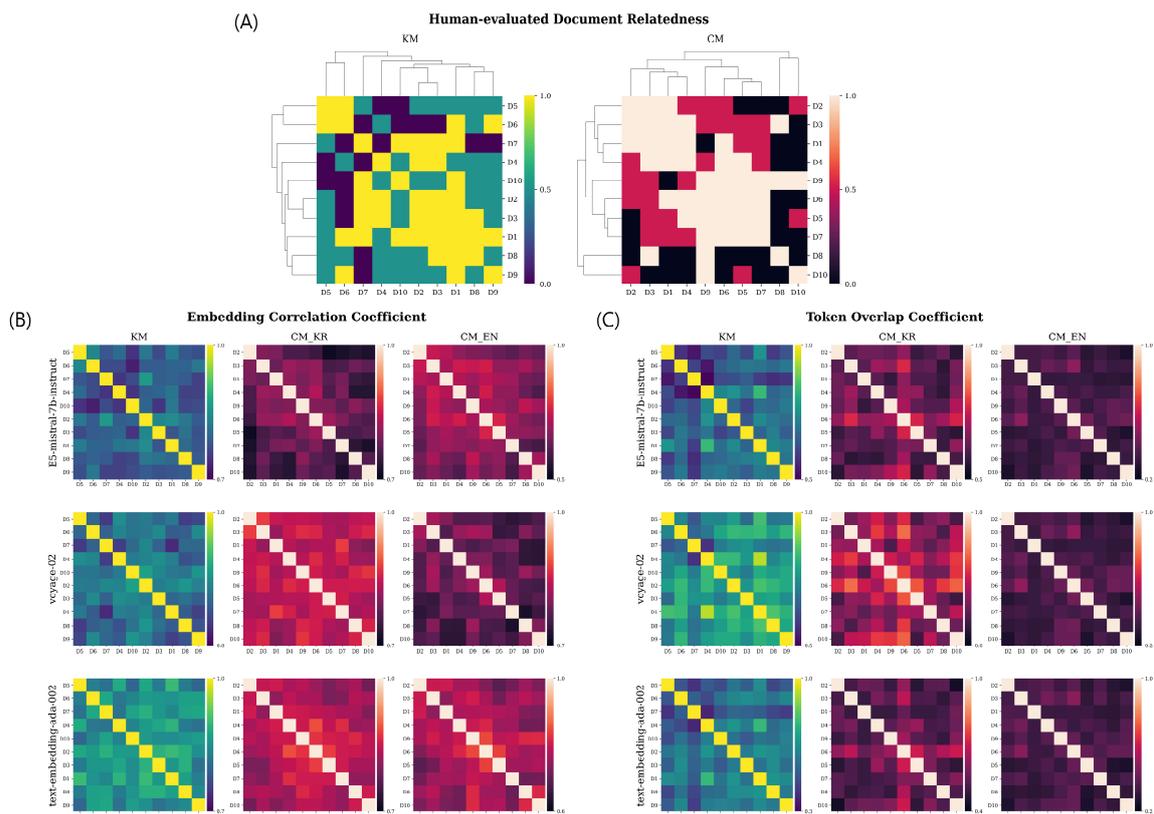

Figure 2. Comparative analysis of human-evaluated document relatedness, embedding correlation coefficients, and token overlap coefficients in KM, CM_KR, and CM_EN. (A) shows clustermaps of human-evaluated document relatedness scores for KM and CM, where each cell represents the perceived relatedness between document pairs as judged by human evaluators. (B) illustrates the embedding correlation coefficients across the different domains and models. (C) depicts the token overlap coefficients, which measure the extent of shared tokens between document pairs. The hierarchical clustering was conducted based on squared Euclidean distance, with embedding correlation coefficients and token overlap coefficients sequentially arranged in an identical order to this cluster structure.

Abbreviations: KM, Korean medicine; CM, Conventional medicine; CM_KR, CM physiology in Korean; CM_EN, CM physiology in English; D, Document.

(2) Correlation analyses between metrics in KM and CM documents

To analyze the correlations between human-evaluated document relatedness scores and embedding correlation coefficients, and between embedding correlation coefficients and token overlap coefficients, Pearson or Spearman correlation coefficients were calculated for each metric pair. Figure 3 provides scatter plots for showing the relationship between the metrics in KM, CM_KR, and CM_EN.

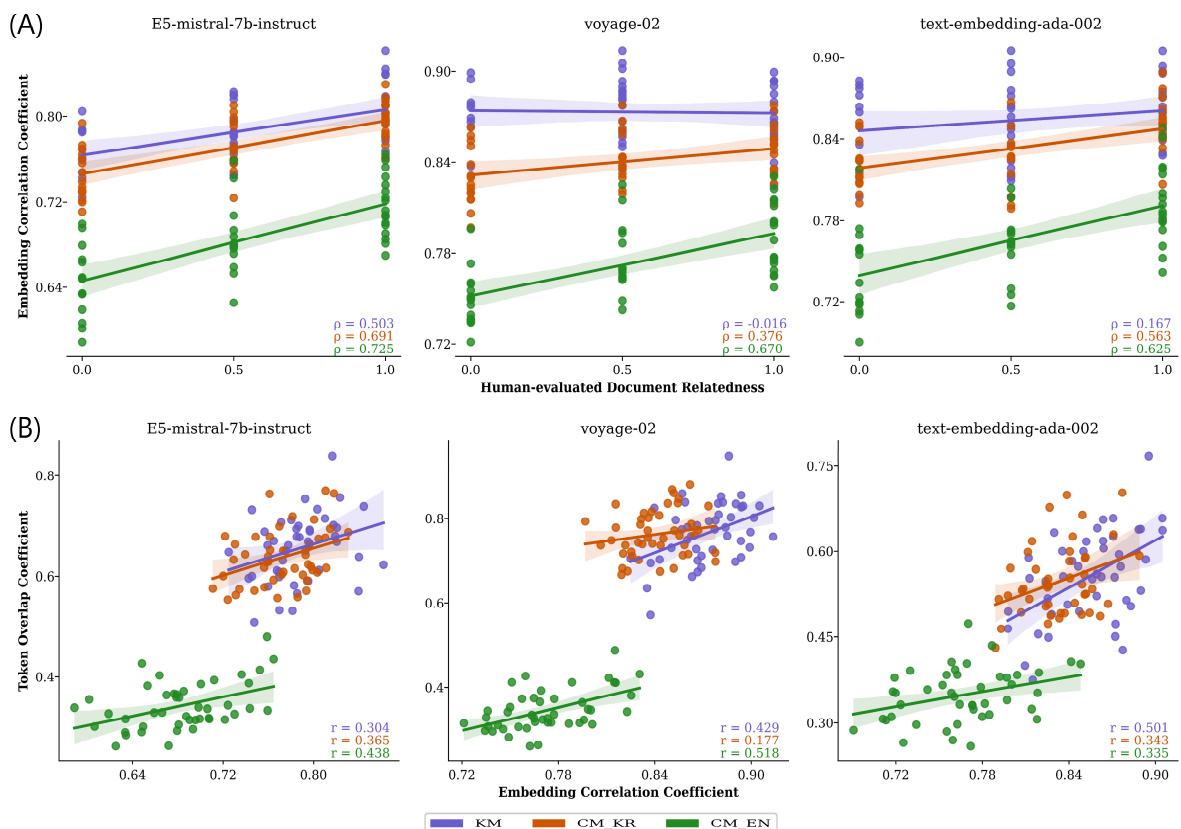

Figure 3. Correlation of document embedding correlation coefficients with human-evaluated document relatedness, and token overlap coefficients in KM, CM_KR, and CM_EN. The figure displays regression plots for pairwise correlations between the metrics within KM, CM_KR, and CM_EN documents. (A) displays scatter plots with fitted regression lines showing the relationship between human-evaluated document relatedness (x-axis) and the embedding correlation coefficient (y-axis) for each of the three language models. Each point represents a document pair. (B) shows the relationship between the embedding correlation coefficients (x-axis) and token overlap coefficients (y-axis). The colors correspond to the different document sets: KM, CM_KR, and CM_EN. The regression lines and correlation coefficients represent the strength and direction of the relationships. The symbols 'r' and 'ρ' indicate the Pearson and Spearman correlation coefficients, respectively.

Abbreviations: KM, Korean medicine; CM, Conventional medicine; CM_KR, CM physiology in Korean; CM_EN, CM physiology in English.

For the first metric pair, Spearman's correlation coefficients were calculated between human-evaluated document relatedness scores and the embedding correlation coefficients. Across all evaluated

models—E5-mistral-7b-instruct, voyage-02, and text-embedding-ada-002—the correlation coefficients for CM were consistently higher than those for KM, indicating a stronger alignment with human judgment in the context of CM. Within CM, the coefficients for CM_EN were higher than those for CM_KR. Specifically, for the E5-mistral-7b-instruct model, the Spearman's correlation coefficient was 0.503 for KM, while it increased for CM_KR to 0.691 and was highest for CM_EN at 0.725. Similarly, voyage-02 presented a negative correlation for KM (-0.016), but it showed positive correlations of 0.376 for CM_KR and a notably stronger 0.670 for CM_EN. The text-embedding-ada-002 model demonstrated a coefficient of 0.167 for KM, with higher values of 0.563 for CM_KR and 0.625 for CM_EN. Notably, CM_EN exhibited statistically significant positive correlations across all models (0.725, 0.670, and 0.625, respectively), indicating a robust positive correlation in the context of CM and English compared to KM and Korean. In contrast, the correlations in KM were either weak or slightly negative (-0.016 and 0.167), with the exception of the E5-mistral-7b-instruct model, which yielded a moderate 0.503.

Secondly, the Pearson correlation coefficients between the embedding correlation coefficients and token overlap coefficients showed varied patterns. In CM_EN, the E5-mistral-7b-instruct model had a Pearson's correlation coefficient of 0.438, and voyage-02 had a coefficient of 0.518, both indicating moderate positive correlations. However, these correlations, including the one for text-embedding-ada-002, were all lower than those observed for human-evaluated document relatedness. For KM, significant positive correlations were observed in voyage-02 and text-embedding-ada-002, with coefficients of 0.429 and 0.501, respectively. These values are in stark contrast to the previously discussed Spearman's correlations between human-evaluated document relatedness scores and embedding correlation coefficients for KM (-0.016 and 0.167, respectively). This suggests that these models may prioritize token-level features of documents over their human-perceived meanings when generating vector representations. These findings are summarized in Table 4.

Table 4. Correlation analysis between document similarity metrics in KM, CM_KR, and CM_EN.

| Embedding model | Human-evaluated document relatedness – Embedding correlation coefficient (Spearman's $\rho$) | | | Embedding correlation coefficient – Token overlap coefficient (Pearson's $r$) | | |
|---|---|---|---|---|---|---|
| | KM | CM_KR | CM_EN | KM | CM_KR | CM_EN |
| **E5-mistral-7b-instruct** | 0.503[b] | 0.691[c] | **0.725[c]** | 0.304 | 0.365 | 0.438[a] |
| **voyage-02** | -0.016 | 0.376 | **0.670[c]** | 0.429[a] | 0.177 | 0.518[b] |
| **text-embedding-ada-002** | 0.167 | 0.563[c] | **0.625[c]** | 0.501[b] | 0.343 | 0.335 |

Superscripts indicate statistical significance in correlation analysis.
[a]p < 0.05, [b]p < 0.005, [c]p < 0.001

Abbreviations: KM, Korean medicine; CM, CM_KR, CM physiology in Korean; CM_EN, CM physiology in English.

Overall, embedding correlations in CM_EN consistently demonstrates a higher alignment with human-evaluated document relatedness compared to KM and CM_KR. On the contrary, the embedding representation of KM tends to be determined by the explicit lexical similarity from token overlaps. These findings illustrate insufficiencies of LLM-based vector embeddings in capturing human-perceived conceptual meanings in niche domains, suggesting that their application in conventional RAG systems may result in suboptimal performances.

2) Performance comparison of Prompt-RAG and existing models
(1) Main results

Table 5 presents the mean scores for relevance, readability, and informativeness, along with the response times for the five models' answers.

Table 5. Comparative evaluation of model performance in the Korean medicine domain

| Model | Relevance (Mean score) | Readability (Mean score) | Informativeness (Mean score) | Response time (Mean seconds) |
|---|---|---|---|---|
| **ChatGPT-3.5** | 1.711 | 1.900 | 0.667[d] | - |
| **ChatGPT-4** | 1.833 | **1.922** | 1.033[b] | - |
| **C50-V300** | 1.733 | 1.733[a] | 0.644[d] | 6.454[d] |
| **C100-V150** | 1.8 | 1.722 | 0.833[d] | 7.033[c] |
| **Prompt-RAG** | **1.956** | 1.900 | **1.589** | 24.840 |

Superscripts indicate statistical significance in comparison to the Prompt-RAG model.
[a]p < 0.05, [b]p < 0.01, [c]p < 0.005, [d]p < 0.001

Firstly, we compared the performance of our prompt-RAG model with that of ChatGPT to examine its proficiency in the KM domain. Prompt-RAG achieved mean scores of 1.956 for relevance and 1.589 for informativeness, respectively, surpassing ChatGPT-3.5 (1.711 for relevance, 0.667 for informativeness) and ChatGPT-4 (1.833 for relevance, 1.033 for informativeness). It is noteworthy that our model's informativeness scores were significantly higher, being more than double those of ChatGPT-3.5 and exceeding those of ChatGPT-4 by over 1.5 times. In terms of readability, our model scored 1.900, which was about equal to ChatGPT-3.5's score (1.900) and slightly lower than ChatGPT-4's (1.922). Overall, our model demonstrated its outperformance against ChatGPT baselines, especially GPT-3.5, in generating domain-specific answers related to KM.

Further, we explored whether the prompt-RAG approach could produce better answers than the conventional chunk retrieval method. For all the criteria, our model scored higher than C50-V300 and C100-V150. The readability scores of our model were significantly higher compared to C100-V150, and especially for informativeness, our model obtained statistically significant scores, approximately 2.5 times that of C50-V300 and around 1.9 times that of C100-V150. However, our mode was significantly slower in terms of average response time, taking an additional 18.356 seconds compared to C50-V300 and 17.806 seconds more than C100-V150. These results find that the Prompt-RAG model excelled in answer quality, while the latency in answer generation was larger than the chunk retrieval method.

(2) Comparison by types of questions

To assess the overall quality and applicability of our prompt-RAG, we conducted a comparative analysis of its performance against the other models across different question types: direct retrieval, comprehensive understanding, and functional robustness. The summed scores for relevance, readability, and informativeness by the three evaluators were averaged for each question and each question type, respectively. The results by the question types are illustrated in Figure 4.

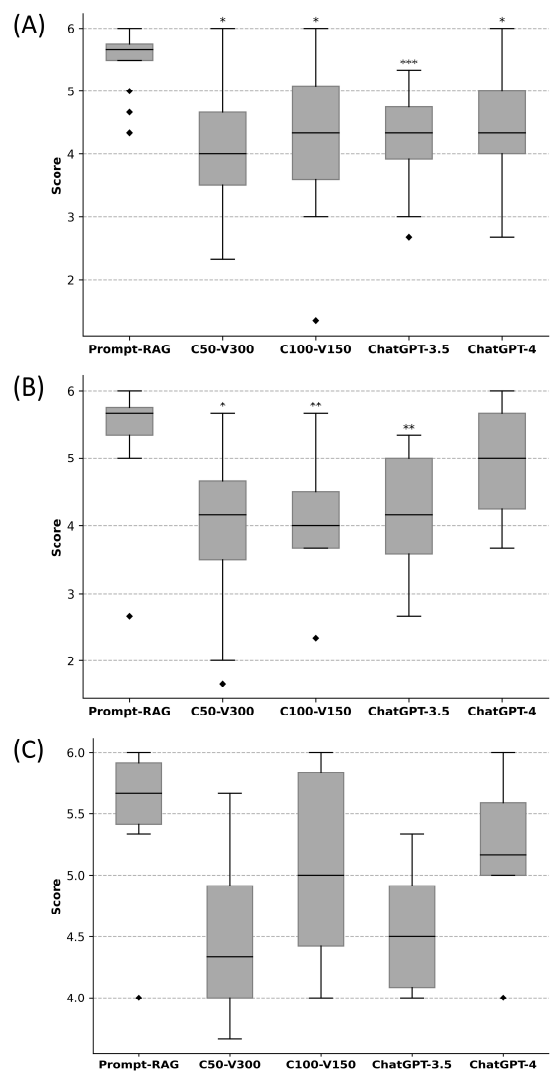

Figure 4. Model performance comparison across different question types. (A) Direct retrieval questions. (B) Comprehensive understanding questions. (C) Functional robustness questions. The asterisks

represent statistical significance in the differences in scores between the prompt-RAG model and the others: *p < 0.05, **p < 0.01, ***p < 0.005

Our model reached an average score of 5.5 for direct retrieval, 5.389 for comprehensive understanding, and 5.444 for functional robustness out of 6, outdoing all other models in every question type. Notably, the scores for direct retrieval were significantly higher compared to those of all the other models, and the scores for comprehensive understanding were also statistically significant in comparison to the chunk retrieval models and ChatGPT-3.5. This suggests not only our model's advanced capability for retrieval but also its comprehension-based answering performance, which is comparable to ChatGPT-4.

## 5. Discussion

In this study, our exploration of LLM-based vector embeddings revealed marked limitations within the KM domain. The analysis showed that vector embeddings are heavily influenced by languages and token overlaps, which are not always compatible with human reasoning, potentially leading to suboptimal performance when used in RAG methods. To address these shortcomings, we introduced Prompt-RAG, a natural language prompt-based RAG methodology, providing a strategic shift from conventional RAGs operated with vector embeddings. This stemmed from the recognition of the limitations inherent in LLMs, utilizing the linguistic capabilities of LLM and addressing its constraints at the same time. As a result, our QA chatbot equipped with Prompt-RAG exhibited promising outcomes in terms of relevance, readability, and informativeness in the KM domain. Moreover, it coped with a variety of types of KM-related questions as well, proving its practical stability.

The potential of Prompt-RAG is substantial. Importantly, our model is not confined only to the KM domain but can be applied to other marginal domains that require RAG. GPT is recognized for its emergent properties, potentially helping deal with highly abstract, contextual, or previously unseen expressions.(37-39) It would facilitate high-quality retrieval with a ToC that contains the comprehensive and essential context of documents, leading to desirable responses across various domains. Its applicability and efficiency can expand vastly, together with natural language processing techniques developing and improving. As the cognitive abilities of LLMs continue to advance, we look forward to Prompt-RAG becoming an even more powerful tool with full reliance on the capabilities of an LLM itself.

Its wide-ranging adaptability derived from the ability to understand and process unacquainted or uncertain concepts and terminologies would raise some challenges for conventional vector embedding-based RAG. For example, a short query has been known to undermine the performance vector embedding-based informational retrieval due to the lack of contexts, even though it is the major form of a search query on the internet.(40-42) The adoption of the natural language prompts through GPT allows for a nuanced understanding of queries(43) and thus results in a more detailed, accurate, and relevant retrieval. In addition, Prompt-RAG can be much more efficient when it comes to model updates, saving on the expense and time for the renewal of document embeddings, especially with larger documents. These properties would be highlighted in dynamic environments in terms of data with its ability to be applied without the need for repetitive retraining or embedding.

However, we acknowledge that Prompt-RAG has certain limitations. Firstly, the requirement for a ToC might sometimes pose an obstacle, depending on the type or structure of the document. Secondly, the recurring latency and expenses associated with running a generative model or making Application Programming Interface (API) calls for heading selection do result in longer response times and higher costs. However, these issues are expected to naturally improve as the generative performance of LLMs continues to develop and model pricing plans become more economical, as has been the trend. Explorations and developments in model compression and light-weight artificial intelligence technologies for resource-constrained devices have been recently encouraged by the popularization of individual edge devices.(44-46) This trend seems to be extending to natural language processing domains as well(47), which would help solve the latency issue of our model. The rapid advancements in generative models suggest that the limitations of our model will become increasingly less problematic in the foreseeable future, likely sooner than anticipated.

## 6. Conclusion

We suggest Prompt-RAG as an alternative to the conventional vector embedding RAG methods, addressing the limitations of LLM-based vector embeddings in niche domains where inconsistencies with human reasoning can lead to suboptimal performance. With its derived QA chatbot, Prompt-RAG has achieved notable outcomes as demonstrated by our study on KM, showing its potential as a versatile and effective tool in line with the rapidly evolving LLM field. While there is room for improvement, its practical benefits are expected to grow through internal and external development. Providing a new paradigm in RAG, it contributes to the advancement of information retrieval in specific domains with remarkable ease.

## 8. Appendix

Table 1. Documents for embedding comparison.

| | **Korean Medicine (KM)** | **Conventional Medicine (CM)** |
|---|---|---|
| **Document 1** | Yin-Yang Perception of Life Phenomena | $Na^+$-$K^+$ ATPase ($Na^+$-$K^+$ Pump) |
| **Document 2** | Six Qi as Analytical Concepts in Life Phenomena: External and Internal Six Qi | Types of Synapses |
| **Document 3** | The Action of Qi | Organization of the nervous system |
| **Document 4** | Physiological Functions of Body Fluids | Circuitry of the cardiovascular system |
| **Document 5** | Analogous Functional System | Erythropoietin |
| **Document 6** | The Concept of Extraordinary Fu Organs | Regulation of Renal Blood Flow |
| **Document 7** | Six Meridians | Acid-Base Disorders |
| **Document 8** | Seven Emotions and Physiological Changes | Satiety |
| **Document 9** | The Concept of Heavenly Water and Menstruation | Negative Feedback Acid-Base Disorders |
| **Document 10** | Sleep and Health Preservation | Pulsatile Secretion of GnRH, FSH, and LH |

The document titles in the Korean Medicine domain are originally in Korean and have been translated for this table.

Table 2. Questions and their types for model evaluation.

| **1. Direct retrieval (40%): 12 Questions** |
| --- |
|   **1) Factual Questions: (1) – (9)** |
|   **2) Comparative Questions: (10) – (12)** |

(1) What is the modernization of Korean medicine (mentioned by the author)[a]?

(2) Can you tell me about Earth from the five elements?

(3) Explain what Congenital Foundation is.

(4) Tell me the constitutional medicine patterns of Taiyin personality.

(5) What are the detailed classifications of sub-health?

(6) What are the new drugs developed based on domestic herbal medicine in Korea?

(7) When is the implementation period for the Fourth Comprehensive Plan for the Promotion and Development of Korean Medicine?

(8) What are the current subjects of the Korean National Licensing Examination for Korean Medicine Doctors?

(9) When was the Law of the People's Republic of China on Traditional Chinese Medicine implemented?

(10) What are the conceptual differences between Blood and Body Fluid?

(11) Compare the classification of the herbs and the formulas.

(12) Can you explain the medical insurance coverage items for Korea, China, and Japan?

| **2. Comprehensive understanding (40%): 12 Questions** |
| --- |
|   **1) Interpretative Questions: (13) – (15)** |
|   **2) Inference Questions: (16) – (18)** |
|   **3) Application Questions: (19) – (21)** |
|   **4) Open-ended Questions: (22) – (24)** |

(13) If you should summarize the meanings of the 'scientification of Korean medicine' into two main points, what would they be?

(14) What aspects contribute to the statement (by the author)[a] that "Korean acupuncture medicine has diversity."?

(15) Tell me about the correlation between Japanese doctors' perceptions of traditional herbal medicine and their actual usage of it.

(16) What is the organ common both in Six Fu and Extraordinary Fu?

(17) Which system of pattern differentiation is most related to the use of Eight Principle pharmacopuncture?

(18) What is the relationship between the pharmacological characteristics of herbal medicine and systems biology?

(19) Patient A has come to a Korean medicine clinic with symptoms of dizziness, tremors, paralysis, convulsions, and itchiness. What exogenous etiological factor seems to cause this?

(20) Patient A received national health insurance coverage for herbal formulas for dysmenorrhea in April of this year. If she visits the clinic for dysmenorrhea in October of the same year, would she be able to receive national health insurance coverage for the herbal formula again?

(21) To become a specialist in internal Korean medicine in 2023, by what year at the latest should one start the general intern program?

(22) Should the use of modern diagnostic medical devices be prohibited in Korean medicine?

(23) What is the significance of the meridian system theory?

(24) What does the future hold for Korean medicine?

**3. Functional Robustness (20%): 6 Questions**
   **1) Adversarial Questions: (25) – (28)**
   **2) Contextual/Reference Questions: (29), (30)**

(25) It is claimed (in the book)[a] that Korean medicine has already been sufficiently modernized and scientized, isn't it?

(26) Triple Energizer is one of Zang-Fu, which is said to be related to the thoracic and abdominal cavities and Qi transformation. Which is more correct?

(27) Is a study where patients are randomly assigned into two groups to test the association between exposure and outcome referred to as a case-control study?

(28) Is it safe to consume ginseng and black goat at the same time?

(29) (Following Question (8)) What are the subjects of the second session of the exam?

(30) (Following Question (16)) Tell me about its physiological functions and the associated Zang-Fu in the context of the Exterior-Interior connection.

[a]This was omitted when the question was posed to ChatGPT.

The questions are originally in Korean and have been translated for this table.